# A Novel Block-DCT and PCA Based Image Perceptual Hashing Algorithm

Zeng Jie[1]

[1] College of Information Engineering, Shenzhen University
Shenzhen, Guangdong, P.R.China
zengjie@szu.edu.cn

**Abstract**
Image perceptual hashing finds applications in content indexing, large-scale image database management, certification and authentication and digital watermarking. We propose a Block-DCT and PCA based image perceptual hash in this article and explore the algorithm in the application of tamper detection. The main idea of the algorithm is to integrate color histogram and DCT coefficients of image blocks as perceptual feature, then to compress perceptual features as inter-feature with PCA, and to threshold to create a robust hash. The robustness and discrimination properties of the proposed algorithm are evaluated in detail. Our algorithms first construct a secondary image, derived from input image by pseudo-randomly extracting features that approximately capture semi-global geometric characteristics. From the secondary image (which does not perceptually resemble the input), we further extract the final features which can be used as a hash value (and can be further suitably quantized). In this paper, we use spectral matrix invariants as embodied by Singular Value Decomposition. Surprisingly, formation of the secondary image turns out be quite important since it not only introduces further robustness, but also enhances the security properties. Indeed, our experiments reveal that our hashing algorithms extract most of the geometric information from the images and hence are robust to severe perturbations (e.g. up to %50 cropping by area with 20 degree rotations) on images while avoiding misclassification. Experimental results show that the proposed image perceptual hash algorithm can effectively address the tamper detection problem with advantageous robustness and discrimination.

*Keywords*: *image hash; perceptual hash; tamper detection*

## 1. Introduction

Image perceptual hashing, also known as image robust hashing, is defined as mapping images to a short bit string following the human perception [1]. In contrast to classic hash functions (MD5, SHA-1), which is highly sensitive to every bit of input data, image perception hashing is sensitive to image content rather than the integrity of image date. The two principal properties of image perception hashing are robustness and discrimination. Robustness means that the hash algorithm should result in the same out bit string for images with the same underlying content. For example, the raw image, its added noise version, its compressed version, its changed brightness version and its rotation angle version have the same underlying content and should share the same hash value. Discrimination implies that the hash values for any two distinct images should be different and random. That is to say, image perceptual hash functions are statistically independent to different image content.

Image perceptual hash value can be used for content identification and digital signature. The former is mainly used in content indexing and analysis, large-scale image database management. The latter is mainly used in the image certification and authentication, digital watermarking. According to the needs of applications, image perceptual hashing should also meet other two properties—randomness and scale-independence. Randomness means that the hash function should withstand all kinds of forgery attack since the hash values are impossible to be reconstructed by the attacker. Scale-independence implies that the length of hash values should always be an even number, although the input images are in different resolution.

Many image perceptual hash functions have been proposed in the literature. Bian Yang[2] uses the mean of image blocks to obtain a perceptual hash. J.Fridrich[3] extracts perceptual features by projecting image blocks onto key based random patterns and thresholds to create a robust hash with the median. R.Venkatesan[4] and M.K. Mihcak[5] selects the low-frequency sub-band of wavelet coefficients to generate a perceptual hash. F.Lefbvre[6] and J.S.Seo[7] uses radon transform to produce a perceptual hash. Hui Zhang[8] creatively introduces the human visual system to obtain a image perceptual hash.

This article addresses the problem of the tamper detection problem of images with image perceptual hashing. Although there are so many image perceptual hash methods proposed, the tradeoff between robustness and discrimination is relatively few discussed. In this article, we aim at proposing a robust and discriminative image perceptual hash algorithm, and explore the algorithm in the application of tamper detection.

Our hashing algorithms consist of two major stages. In the first stage we derive robust feature vectors from pseudo-

This work was supported by Natural Science Foundation of China (61103174)，Science and Technology Program of Shenzhen(JC201105170647A)，Laboratory and Device Management Research Foundation of Shenzhen University(2011045).

random (PR) semi-global regions via matrix invariants; these features are termed "intermediate features". In the second stage, the intermediate features are used to construct a PR secondary image, which is then used to extract the final feature vectors, which constitute the hash value of the image. We observe that the second stage increases robustness against attacks (e.g., compression, rotation, cropping, etc.) and conjecture that it limits information leakage to the adversary. By extracting the hash value from the secondary image, we effectively reduce the effects of the geometric attacks. Recall that, the idea of extracting features from pseudo-randomly selected regions for hashing has been considered before [3, 4, 5, 6]. A similar approach was followed in [7] to design an image watermarking algorithm. In contrast with the prior robust image hashing work, in this paper we employ PCA components to capture essential characteristics of images rather than PR linear statistics. In summary, the main contribution of this work is to view PR image portions as linear operator representatives (i.e., matrices) and to use matrix invariants on them, in particular PCA, to extract robust PR features and to employ them to derive robust hash values. Thus, we effectively propose a new PR signal representation and illustrate its usage within the context of the image hashing problem.

The rest of this paper is organized as follows. In Section 2, we give a general description of the image hashing problem and then give an algorithmic description of our hashing approach. In the meanwhile, we continue with our candidate hashing algorithms, where we give a brief description of each one. Section 3 describes the proposed robust and discriminative image perceptual hash algorithm. The experimental results are detailed and we test our algorithms under various attacks and report the performance results in Section 4. The properties of the hash value are also contained in the Section 4. Conclusion and future work are introduced in Section 5.

## 2. HASHING ALGORITHMS

A robust image hash function for security purposes has two inputs, an image $I$ and a secret key $k$ and produces a short binary vector $\vec{h} = H_k(I)$ from the set $\{0,1\}^h$. The hash function should possess perceptual properties: Hash values for all perceptually "approximately-the-same" images are desired to be equal with high probability; in contrast, perceptually different images should produce independent hash values with high probability. Obviously, such a hash function is a many-to-one mapping. The notion of acceptable disturbances is not precise, and in this version it can be taken to mean that we consider two images to be similar when there are unnoticeable visual distortions between them in terms of human perception. We will address this issue in our future work. We formulate these requirements as follows:

1) Randomization: For any given input $I$, its hash value should be approximately uniformly distributed among all possible $2^h$ outputs:
$$\forall h \in \{0,1\}^h, \Pr\{H_k(I) = \vec{h}\} \approx 2^{-h}$$

2) Pairwise Independence: The hash outputs for two perceptually different images (say $I_1$ and $I_2$) should be approximately independent:
$$\forall h_1, h_2 \in \{0,1\}^h, \Pr\{H_k(I_1) = \vec{h}_1 H_k(I_2) = \vec{h}_2\}$$
$$\approx \Pr\{H_k(I_1) = \vec{h}_1\}$$

3) Invariance: For all possible acceptable disturbances, the output of the hash function should remain approximately invariant. Let $I$ and $\hat{I}$ be perceptually similar images. Then,
$$\forall h \in \{0,1\}^h, \Pr\{H_k(I) = \vec{h}\} \approx \Pr\{H_k(\hat{I}) = \vec{h}\}$$

Next, we present the algorithmic description of our generic hashing scheme:

**Step 1**: Let the $n \times n$ input image be $I \in R^{n \times n}$

**Step 2**: From $I$, pseudo-randomly form $p$ possibly overlapping rectangles (each of them of size $m \times m$): $A_i \in R^{m \times m}, 1 \leq i \leq p$.

**Step 3**: Generate a feature vector $\vec{g}_i$ from each rectangle $\vec{A}_i$ via the transformation $\vec{g}_i = T_1(A_i)$.

**Step 4**: Construct a secondary image $J$ S by using a $PR$ combination of intermediate feature vectors $\{\vec{g}_1, ..., \vec{g}_p\}$.

**Step 5**: From $J$, pseudo-randomly form $r$ possibly overlapping rectangles (each of them of size $d \times d$): $B^i \in R^{d \times d}, 1 \leq i \leq r$.

**Step 6**: Generate a final feature vector $\vec{f}_i$ from each rectangle $B^i$ via the transformation $\vec{f}_i = T_2(B_i)$

**Step 7:** Combine $\{\vec{f}_1, ..., \vec{f}_r\}$ to form the final hash vector.

The choice of the transformations $T_1$ and $T_2$ is crucial for the performance of the scheme; we propose to use PCA for $T_1$, or $T_2$ or both. In principle, other approximately-invariant matrix decompositions may also be employed. Furthermore, the formation of these secondary images $J$ improves the results considerably. In each step of the algorithm, pseudo-randomization is achieved via a secure PR number generator by using the same key, or part of a common key as the seed; this key is unknown to the attacker. Note that, in this paper, we particularly focus on robust PR feature generation from images; in the formation of the final hash values, further dimensionality reduction can, in principle, be achieved via quantization with $PR$ lattices or

PR projection to lower-dimensional appropriately-chosen subspaces.

In general, the choice of transformations is a significant and nontrivial task: We would like to capture the essence of the geometric information while having dimensionality reduction and introducing enough randomness. In case of conventional transforms (such as DCT or DWT, which have proven to be effective for traditional applications), the image is projected onto a fixed set of basis vectors. It is still an open question, however, which mappings (if any) from DCT/DWT coefficients preserve the essential information about an image for hashing and/or mark embedding applications. In addition, from a security point of view, it may be beneficial if the basis vectors of the transform of interest are pseudo-randomly adapted to the image to minimize information leakage to an adversary.

Unlike DCT/DWT-type fixed basis transforms, PCA selects the optimal basis vectors in $L_2$ norm sense such that, for any $m \times m$ real matrix $A, \forall k, 1 \leq k \leq m$, we have

$$(\sigma_k, u_k, v_k) = \arg\min_{a,\vec{x},\vec{y}} \left| A - \sum_{l}^{k-1} \sigma \vec{u}_l \vec{v}_l^T - a\vec{x}\vec{y}^T \right|_F^2$$

Where $a \in R, \vec{x}, \vec{y} \in R^m, \sigma_1 \geq \sigma_2 \geq ... \geq \sigma_m$ are the singular values, $u_i$ and $v_i$ are the corresponding singular vectors and $(\cdot)^T$ is the transpose operator. As an analogy, if an image is represented as a vector in some high-dimensional vector space, then the singular vectors give the optimal "directional information" about the image in the sense of the above formula, while the singular values give the distance information along this "direction". Consequently, the singular vectors that correspond to large singular vectors are naturally prone to any scaling attack and other small conventional signal processing modifications.

By using PCA, we view an image as a two dimensional surface in a three dimensional space. When DCT-like transformations are applied to an image (or surface), the information about any particularly distinctive (hence important) geometric feature of the image is dispersed to all coefficients. As an example, a surface with strong peaks (e.g., very bright patches in a dark background) will be dispersed to all transform in case of DCT. By using PCA, we preserve both the magnitude of these important features (in singular values) and also their location and geometry in the singular vectors. Hence, the combination of the "top" left and right singular vectors (i.e., the ones that correspond to the largest singular values) capture the important geometric features in an image. We utilize this observation in our algorithms.

As a variant of our previous approach, we first use 2D-DCT as the initial transform (i.e., $T_1$) in step 3 of the generic hashing algorithm; which yields the DCT-PCA method. Similar to DCT-PCA, a DWT-PCA method can be derived using the DWT transformation instead of the DCT in step 3 of the generic algorithm. Due to space limitations, here we only focus on DCT-PCA method and report that DWT-PCA approach yields similar results experimentally.

Let $D_i$ be the 2D-DCT of each sub-image $A_i, 1 \leq i \leq p$. After computing $\{D_i\}_{i=1}^{p}$, we keep the coefficients that correspond to low-to-mid band frequencies from each $D_i$.

Although we believe that there is no particular frequency band of DCT coefficients that can be clearly considered as more important than the others, we observed that the coefficients of low-to-mid band frequencies carry more descriptive and distinctive information about images. We avoid near DC frequencies which are more sensitive to simple scaling or DC level changes. By selecting a relatively small value of band frequency, we avoid using coefficients of higher frequencies, which can be altered significantly by noise addition, smoothing, compression, etc. In general, depending on the problem specifications, suitable values of band frequency can be chosen.

The coefficients in this frequency band are then stored as a vector $\vec{d}_i \in R^{f_{max} \times f_{max} - f_{min} \times f_{min}}$ for each rectangle $A_i$. The ordering of the elements of $\{\vec{d}_i\}$ is user dependent and can possibly be used to introduce extra randomness. Then, we proceed with a similar method to the one given in Sec. 2. First we form the set $\Gamma\{\vec{d}_1,...,\vec{d}_p\}$ and then form the PR smooth secondary image $J$, the method of formation of $J$ given $\Gamma$ is the same as the one explained n Sec. 2. Next, we find the PCA of the secondary image $J : J = USV^T$ and store the first left and right singular vectors $\vec{u}_1$ and $\vec{v}_1$ as the hash value of the image, i.e., $h = \{\vec{u}_1, \vec{v}_1\}$.

## 3. Robust and Discriminative Image Perceptual Hashing

The main idea of the algorithm is to integrate color histogram and low-frequency Discrete Cosine Transform (DCT) coefficients of image blocks as perceptual features, then to compress perceptual features as inter-features with Principal Component Analysis (PCA), and to threshold to create a robust hash. The framework of this algorithm is shown in Figure 1.

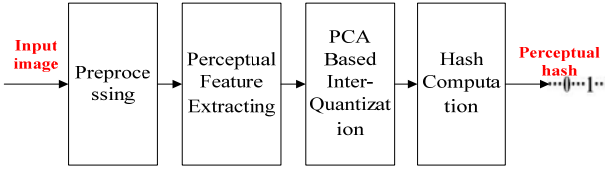

Figure 1. Flow chart of our image perceptual hash method

### 3.1 Image Preprocessing

The research of cognitive psychology and human visual system show that the sensitivity of eyes to chroma signal is much weaker than to luminance signal, and that brightness is the main features of the image signal[1]. So, only the luminance information is considered in preprocessing. The input image is first converted to a standardized image (64*64) via re-sampling and interpolation.

Preprocessing not only reduces the computational complexity of follow-up steps (perceptual feature extracting, PCA based inter-quantization, hash computation), but also ensures that the algorithm is independent of scale.

### 3.2 Perceptual Feature Extracting

During perceptual feature extracting, we adopt the block images strategy, and integrate color histogram and low-frequency Discrete Cosine Transform coefficients of every image block as perceptual features. The process is detailed as follows:

- Divide the standardized image into 64 blocks (block size: 8*8).
- Calculate the color histogram of blocks successively, and the calculation formula is as follows:

$$hist(i) = count(|ima_{gray}/32|), \quad i=0,1,\cdots,7 \qquad (1)$$

- Select the DCT coefficients (DC coefficient and 7 AC coefficients) of blocks successively, and integrate the color histogram as perceptual feature, shown in figure 2.

The energy of the image will be gathered into some DCT coefficients after DCT transformation, DC coefficients contain the main information of the original data matrix, AC coefficients contain the detail information of the data matrix. Meanwhile, they are the most sensitive information of human visual system.

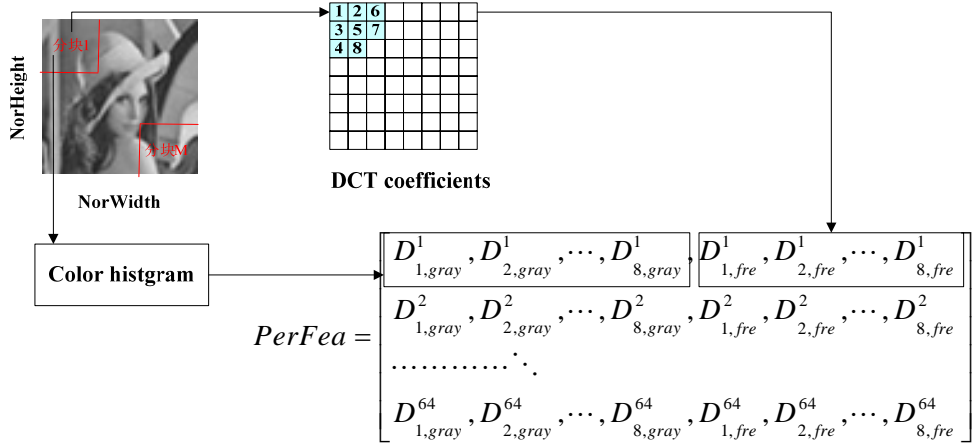

$$PerFea = \begin{bmatrix} D^1_{1,gray}, D^1_{2,gray}, \cdots, D^1_{8,gray}, D^1_{1,fre}, D^1_{2,fre}, \cdots, D^1_{8,fre} \\ D^2_{1,gray}, D^2_{2,gray}, \cdots, D^2_{8,gray}, D^2_{1,fre}, D^2_{2,fre}, \cdots, D^2_{8,fre} \\ \cdots\cdots\cdots\cdots \ddots \\ D^{64}_{1,gray}, D^{64}_{2,gray}, \cdots, D^{64}_{8,gray}, D^{64}_{1,fre}, D^{64}_{2,fre}, \cdots, D^{64}_{8,fre} \end{bmatrix}$$

Figure 2. Process of perceptual feature extracting

### 3.3 PCA Based Inter-Quantization

Each column of perceptual feature (matrix) is an indicator, reflecting appropriate information of the input image. For example, the first column is constituted with each block's occurrences of pixels between 0 and 31, reflecting the distribution of the input image on the pixel interval. However, there is inherent correlation among adjacent blocks. So, each column of perceptual feature matrix contains some redundant information.

Principal Component Analysis is used to reduce, even elimilate, the redundant information in perceptual feature matrix. Then, the perceptual feature is compressed into a inter-feature matrix, with a smaller dimension (10*64) and few redundant information.

### 3.4 Hash Computation

Once the inter-feature matrix is generated, we can obtain the perceptual hash of input image via binarizing. Each column is binarized using the median of the rank-ordered coefficients. If the subset of rank-ordered coefficients is denoted as $c(i), \quad i=0,1,\cdots,K$, then their median is calculated as $\mu = (c(K/2) + c((K+1)/2))/2$. Then, the perceptual hash of input image is obtained by thresholding each column with the median $\mu$ as follows:

$$hash_i = \begin{cases} 1, & c(i) \geq \mu \\ 0, & c(i) < \mu \end{cases}, \quad i=0,1,\cdots,K \qquad (2)$$

## 4. Experimental Results

To asses the performance of each method, we apply them both to standard test images and an image database of about 5000 images. Here, we present some preliminary results under compression, rotation and cropping attacks. We note that the practical choice of algorithmic parameters can further be optimized in order to improve the results. In experiments involving standard test images, we consider the $512 \times 512$ gray scale Lena image and compare its hash value with those of the attacked Lena and Goldhill images. In all simulations, $h_1, h_2, h_3$ denote the hash values of the original Lena, attacked Lena and Goldhill respectively. Although we experimented with a wide range of attacks, including benchmark attacks, here we report results for only a few classes of attacks for the purposes of illustration.

We start our experiments with the algorithmic parameters are chosen as $p = 200, m = 256, r = 200, d = 150$ ; the secondary image $J$ is of size $256 \times 400$. As an attack, we crop 50 percent of the image by area, rotate it 20 degrees and JPEG compress it with quality factor (QF) 5.

The proposed algorithm has been implemented with matlab scripts. The evaluation was based on 72 distinct images. These images are all from corel image galley (URL: http://calphotos.berkeley.edu/).

The bit error rate(BER) [9] [10] is denoted as the rate of mismatched bits by comparing two perceptual hashes.

### 4.1 Robustness to Image Operation

Robustness implies that perceptual hash functions should be robust to all kinds of image operation (contrast increase, median filter, JPEG compression, noise addition, histogram equalisation, laplace sharpen, rotation), since the underlying content is never changed. That's to say, the BER between the perceptual hash of the raw image and the perceptual hash of the operated image should be infinitely close to 0.The experimental results of robustness evaluation is presented in table 1.

TABLE I.   RESULTS OF ROBUSTNESS EVALUATION

| Image Operation | | Mean of BER |
|---|---|---|
| Contrast Increase | -30% | 8.50% |
| | -20% | 3.27% |
| | 20% | 4.10% |
| | 30% | 9.20% |
| Median filter | | 10.06% |
| JPEG Compression | 10% | 3.00% |
| | 20% | 8.22% |
| | 40% | 17.50% |
| Noise Addition | Gaussian | 7.10% |
| | Peper and Salt | 8.50% |
| Histogram Equalisation | | 11.27% |
| Laplace Sharpen | | 20.24% |
| Rotation | -5 | 23.39% |
| | -3 | 15.70% |
| | +3 | 17.33% |
| | +5 | 22.00% |

Table 1 shows the mean of BER between the raw image's perceptual hash and the operated image's perceptual hash. Notice that most of the results are lower than 0.1 and stay close to the theoretical value 0, only when the operation (JGEP compression 40%, rotation 5) has changed the underlying content. Meanwhile, the more the underlying content changes, the bigger the mean of BER is. For example, the result is 0.0300 while the JGEP compression is 10%, and 0.1750 while the JGEP compression is 40%.

### 4.2 Discrimination to Different Images

Discrimination means that the hash functions are statistically independent for different perceptual content, so that any two distinct images result in different and apparently random perceptual hash.

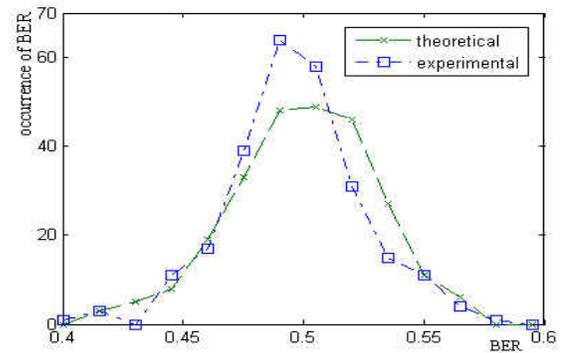

Figure 3.   Distribution of BER between distinct images

Assume that two distinct images ( $i$ and $i'$ ) are perceptually different, the theoretical optimal value of their BER $M_{BER}(pHash)$ can be estimated as follows:

$$M_{BER}(pHash) = E[BER(i,i')] \qquad (3)$$

where $i$ and $i'$ are taken independently and randomly from a given image set, and $E[]$ denotes mathematical expectation. According to Baris Coskun's and Kevin Hamon's analysis and proof in article [9] and [10], the theoretical optimal value is speculated to be 0.5.

Without loss of generality, we calculate BER between 72 test images, using 255 BER computation overall. Then we count the occurrence of each BER value and obtain the experimental distribution of BER, as shown in Figure 3. Meanwhile, we also plot the theoretical probability density function with $\mu = 0.5$ and $\sigma^2 = 0.0009$ in Figure 3.

The BER between perceptual hash of distinct images has a Gaussian distribution around the mean value of 0.4996, which is close to the theoretical optimal value 0.5. Thus, the perceptual hash of different images can be regarded as statistically independent as expectation.

### 4.3 Detection to the Tamper—Addition of a Logo

Tamper always brings in malicious changes to the original content of raw image. Typical image tampering operations include adding LOGO, image mosaic and so on. In this article, we mainly concern the tamper of adding LOGO, since the addition of a LOGO causes minimal change and is widely used with the increasing spread of Internet. The sample of adding LOGO is shown in Figure 4:

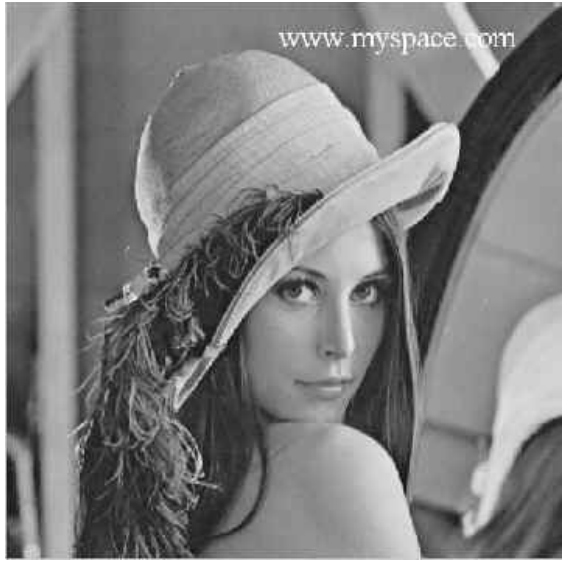

Figure 4. Sample of adding a LOGO

The content of tampered image shown in Figure 4 is much similar with the law image, only with malicious changes via adding a logo on the left. In order to detect such tampering operation to image, the BER between raw image and tampered image should be bigger than the BER of robust operation. Meanwhile, in order to distinguish tampered image from distinct image, the BER between raw image and tampered image should also be smaller than the BER of distinct images. We calculate 12 BER between raw image and tampered image, as shown in Figure 5:

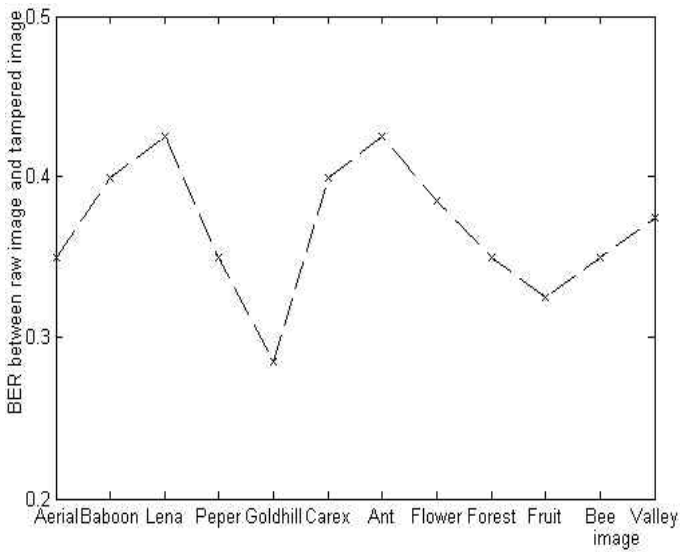

Figure 5. BER between raw image and tampered image

In Figure 5, most of the BER between raw image and tampered image is in the range of [0.3, 0.45]. Note that the most BER of robust operation is lower than 0.1 and the BER of distinct images has a Gaussian distribution around the mean value of 0.4996. Thus, the proposed image perceptual hash algorithm can effectively address such tamper detection problem with advantageous robustness and discrimination.

## 5. Conclusions

In this article we proposed a robust and discriminative image perceptual hash algorithm in order to address the problem of the tamper detection problem of images. We integrate color histogram and low-frequency DCT coefficients of image blocks as perceptual feature, then compress perceptual feature as inter-feature with PCA, and threshold the inter-feature to create a robust hash. Experimental results show that the proposed algorithm is advantageous at robustness since the most BER of robust operation is lower than 0.1, but it is a pity that the robustness toward rotation is not so perfect as assumed. It's also found that the proposed algorithm has a very discriminative power because the BER of distinct images has a Gaussian distribution around the mean value of 0.4996, which is close to the theoretical optimal value 0.5. With such advantageous robustness and discrimination, the proposed algorithm can effectively detect the tampering operation of adding a LOGO. Future investigation will address the problem of verification of the tamper detection ability toward other tampering operations. Meanwhile, we will extend this still-image perceptual hash method to video clip to address the problem of video authentication and copyright protection.

**Acknowledgments**

This work was partially funded by Natural Science Foundation of China through the Free Application Program under contract 61103174, Science and Technology Program of Shenzhen (China) through the Basic Research Program under contract JC201105170647A, and Laboratory and Device Management Research Foundation of Shenzhen University through the Basic Research Program under contract 2011045.

The Authors would thanks to Natural Science Foundation of China and Shenzhen Municipal Science and Technology Trade and Industry and Information Technology Commission (China) for their funding of our projects. The Authors also express gratitude to Laboratory and Device Management Department of Shenzhen University for their support and collaboration.

## References

[1] NIU Xia-mu ,JIAO Yu-hua, "An Overview of Perceptual Ha shing," ACTA ELECTRONICA SINICA.China, Beijing, Vol.36, No. 7, pp. 1405-1411,July, 2008.

[2] Bian Yang, Fan Gu, Xiamu Niu, "Block Mean Value Based Image Perceptual Hashing," International Conference on Intelligent

Information Hiding and Multimedia Signal Processing(IIH-MSP '06), Dec.2006, pp.167-172.

[3] J. Fridrich, "Visual hash for oblivious watermarking," Proc .IS&T/SPIE 12th Annu. Symp., Electronic Imaging, Security and Watermarking of Multimedia Content II, San Jose, CA, Jan. 2000, pp.286-294.

[4] R. Venkatesan, S.Koon, M. Jakubowski, and P. Moulin, "Robust image hashing," Proc. IEEE Int. Conf. Image Processing 2000, vol. 3, pp.664-666.

[5] M. K. MIhCak,R Venkatesan, "New Iterative Geometric Methods for Robust Perceptual Image Hashing," Proc of ACM Workshop on Security and Privacy in Digital Rights Management . Philadelphia :LNCS ,2001,pp.13-21.

[6] F.Lefbvre,B.Macq,J.-D.Legat, "RASH: RAdon Soft Hash algorithm," Proc. EUSIPCO,Toulouse,France,2002. pp.54-61.

[7] J.S.Seo, J.Haitsma, T.Kalker,C.D.Yoo,"A robust image fingerprinting system using the radon transform," Signal Process.: Image Commun., vol. 19, no. 4, 2004,pp. 325–339.

[8] Hui Zhang,Haibin Zhang,Qiong Li, Xiamu Niu, "A Multi-Channel Combination Method of Image Perceptual Hashing," Fourth International Conference on Networked Computing and Advanced Information Management,2008. NCM '08.Volume 2,Sep. 2008,pp.87-90.

[9] Baris Coskun, Bulent Sankur, Nasir Memon, "Spatio-Temporal Transform Based Video Hashing," IEEE Transactions on Multimedia,VOL. 8,NO. 6,Dec .2006. pp. 1190-1208.

[10] Kevin Hamon, Martin Schmuker, Xuebing Zhou, "Histogram-based perceptual hashing for minimally changing video sequence," The Second IEEE International Conference on Automated Production of Cross Media Content for Multi-Channel Distribution 2006.

**Zeng Jie** received his M. Eng. Degree in Signal and Information Processing from the Tianjin University, Tianjin, China in 2001. From 2001 to 2006, He worked as a software engineer in Huawei Technology Ltd. Now he is a lecturer in the College of Information Engineering, Shenzhen University, China. His current research intersts include wireless networks, wireless commmunication, and cooperative wireless networks.